\documentclass[runningheads]{llncs}
\usepackage{graphicx}
\usepackage{amsmath,amssymb} 
\usepackage{color}
\usepackage[width=122mm,left=12mm,paperwidth=146mm,height=193mm,top=12mm,paperheight=217mm]{geometry}

\usepackage{hyperref}
\usepackage{caption, subcaption}
\usepackage{setspace}

\DeclareMathOperator{\argmax}{argmax}
\DeclareMathOperator{\softmax}{softmax}

\newtheorem{thm}{Theorem}
\newtheorem{cor}[thm]{Corollary}
\newtheorem{lem}[thm]{Lemma}
\newtheorem{prop}[thm]{Proposition}

\usepackage{color}

\begin{document}
\pagestyle{headings}
\mainmatter

\title{Improving DNN Robustness to Adversarial Attacks using Jacobian Regularization}

\titlerunning{ECCV 2018 Conference Paper}

\authorrunning{D. Jakubovitz, R. Giryes}

\author{Daniel Jakubovitz, Raja Giryes}
\institute{School of Electrical Engineering,\\
	Tel Aviv University, Israel\\
	\email{danielshaij@mail.tau.ac.il, raja@tauex.tau.ac.il}
}
\maketitle

\begin{abstract}
Deep neural networks have lately shown tremendous performance in various applications including vision and speech processing tasks. However, alongside their ability to perform these tasks with such high accuracy, it has been shown that they are highly susceptible to \emph{adversarial attacks}: a small change in the input would cause the network to err with high confidence. This phenomenon exposes an inherent fault in these networks and their ability to generalize well.
For this reason, providing robustness to adversarial attacks is an important challenge in networks training, which has led to extensive research.
In this work, we suggest a theoretically inspired novel approach to improve the networks' robustness. Our method applies regularization using the Frobenius norm of the Jacobian of the network, which is applied as post-processing, after regular training has finished. We demonstrate empirically that it leads to enhanced robustness results with a minimal change in the original network's accuracy.
\footnote{The code is available at \url{https://github.com/danieljakubovitz/Jacobian_Regularization}}

\end{abstract}

\section{Introduction}
Deep neural networks (DNNs) are a widespread machine learning technique, which has shown state-of-the-art performance in many domains  such as natural language processing, computer vision and speech processing \cite{Goodfellow16Deep}. Alongside their outstanding performance, deep neural networks have recently been shown to be vulnerable to a specific kind of attacks, most commonly referred to as {\em Adversarial Attacks}. These cause significant failures in the networks' performance by performing just minor changes in the input data that are barely noticeable by a human observer and are not expected to change the prediction \cite{Goodfellow15Explaining}. These attacks pose a possible obstacle for mass deployment of systems relying on deep learning in sensitive fields such as security or autonomous driving, and expose an inherent weakness in their reliability.

In adversarial attacks, very small perturbations in the network's input data are performed, which lead to classifying an input erroneously  with a high confidence. Even though these small changes in the input cause the model to err with high probability, they are unnoticeable to the human eye in most cases. In addition, it has been shown in \cite{Intriguing} that such adversarial attacks tend to generalize well across models. This transferability trait only increases the possible susceptibility to attacks since an attacker might not need to know the structure of the specific attacked network in order to fool it. Thus, black-box attacks are highly successful as well. This inherent vulnerability of DNNs is somewhat counter intuitive since it exposes a fault in the model's ability to generalize well in very particular cases.

Lately, this phenomenon has been the focus of substantial research, which has focused on effective attack methods, defense methods and theoretical explanations to this inherent vulnerability of the model.
Attack methods aim to alter the network's input data in order to deliberately cause it to fail in its task. Such methods include  DeepFool \cite{DeepFool}, Fast Gradient Sign Method (FGSM) \cite{Goodfellow15Explaining}, Jacobian-based Saliency Map Attack (JSMA) \cite{JSMA}, Universal Perturbations \cite{Dezfooli17Universal}, Adversarial Transformation Networks \cite{Baluja18Learning}, and more \cite{Carlini17Towards}.

Several defense methods have been suggested to increase deep neural networks' robustness to adversarial attacks. Some of the strategies aim at detecting whether an input image is adversarial or not (e.g., \cite{Metzen17detecting,Li17Adversarial,Lu17SafetyNet,Xu18Feature,Meng17MagNet,Detecting}). For example, the authors in \cite{Xu18Feature} suggested to detect adversarial examples using feature squeezing, whereas the authors in \cite{Detecting} proposed to detect adversarial examples based on density estimates and Bayesian uncertainty estimates.
Other strategies focus on making the network more robust to perturbed inputs. The latter, which is the focus of this work, aims at increasing the network's accuracy in performing its original task even when it is being fed with perturbed data, intended to mislead it. This increased model robustness has been shown to be achieved by several different methods. 

These defense methods include, among others, Adversarial Training \cite{Goodfellow15Explaining} which adds perturbed inputs along with their correct labels to the training dataset; Defensive Distillation \cite{Distillation}, which trains two networks, where the first is a standard classification network and the second is trained to achieve an output similar to the first network in all classes; the Batch Adjusted Network Gradients (BANG) method \cite{BANG}, which balances gradients in the training batch by scaling up those that have lower magnitudes; Parseval Networks \cite{Parseval} which constrain the Lipschitz constant of each hidden layer in a DNN to be smaller than 1; the Ensemble method \cite{Ensembling}, which takes the label that maximizes the average of the output probabilities of the classifiers in the ensemble as the predicted label; a Robust Optimization Framework \cite{RobustOptimization}, which uses an alternating minimization-maximization procedure in which the loss of the network is minimized over perturbed examples that are generated at each parameter update; Virtual Adversarial Training (VAT) \cite{Virtual_Adversarial_Training}, which uses a regularization term to promote the smoothness of the model distribution; Input Gradient Regularization \cite{InputGradients} which regularizes the gradient of the cross-entropy loss, and Cross-Lipschitz Regularization \cite{Cross_Lipschitz_regularization}, which regularizes all the combinations of differences of the gradients of a network's output w.r.t its input. In another recent work \cite{CertifyingRobustness}, the authors suggested an adversarial training procedure that achieves robustness with guarantees on its statistical performance.

In addition to these works, several theoretical explanations for adversarial examples have been suggested. 
In \cite{Goodfellow15Explaining}, the authors claim that linear behavior in high-dimensional spaces creates this inherent vulnerability to adversarial examples.
In \cite{Oh17Adversarial}, a game theoretical framework is used to study the relationship between attack and defense strategies in recognition systems in the context of adversarial attacks.
In \cite{Transferable}, the authors examine the transferability of adversarial examples between different models and find that adversarial examples span a contiguous subspace of large dimensionality. The authors also provide an insight into the decision boundaries of DNNs. 
In \cite{Towards_DeepLearning_Resistance}, the authors claim that first order attacks are universal and suggest the Projected Gradient Descent (PGD) attack which relies on this notion. They also claim that networks require a significantly larger capacity in order to be more robust to adversarial attacks.
In another recent work \cite{Vulnerability_Input_Dimension}, the authors show that the gradient of a network's objective function grows with the dimension of its input and conclude that the adversarial vulnerability of a network increases with the dimension of its input.

In \cite{ClassificationRegions}, the authors showed the relationship between a network's sensitivity to additive adversarial perturbations and the curvature of the classification boundaries. In addition, they propose a method to discriminate between the original input and perturbed inputs.
In \cite{Curvature}, the link between a network's robustness to adversarial perturbations and the geometry of the decision boundaries of this network is further developed. Specifically, it is shown that when the decision boundary is positively curved, small universal perturbations are more likely to fool the classifier. However, a direct application of this insight to increase the networks' robustness to adversarial examples is, to the best of our knowledge, still unclear.

In a recent work \cite{Sokolic17Robust}, a relationship between the norm of the Jacobian of the network and its generalization error has been drawn. The authors have shown that by regularizing the Frobenius norm of the Jacobian matrix of the network's classification function, a lower generalization error is achieved.
In \cite{GradientRegularization} the authors show that using the Jacobian matrix computed at the logits (before the softmax operation) instead of the probabilities (after the softmax operation) yields better generalization results.

Inspired by the work in \cite{Sokolic17Robust}, we take this notion further and show that using Jacobian regularization as post-processing, i.e. applying it for a second phase of additional training after regular training has finished, also increases deep neural networks' robustness to adversarial perturbations. Besides the relationship to the generalization error, we show also that the Forbenius norm of the Jacobian at a given point is related to its distance to the closest adversarial example and to the curvature of the network's decision boundaries. 
All these connections provide a theoretical justification to the usage of the Jacobian regularization for decreasing the vulnerability of a network to adversarial attacks.

We apply the Jacobian regularization as post-processing  to the regular training, after the network is stabilized with a high test accuracy, thereby allowing to use our strategy with  existing pre-trained networks and improve their robustness. In addition, using the Jacobian regularization requires only little additional computational resources as it makes a single additional back-propagation step in each training step, as opposed to other methods that are very computationally demanding such as Distillation \cite{Distillation} which requires the training of two networks.

Two close techniques to our strategy are the Input Gradient regularization technique proposed in \cite{InputGradients} and the Cross-Lipschitz regularization proposed in \cite{Cross_Lipschitz_regularization}.
Our approach differs from the former work by the fact that we regularize the Frobenius norm of the Jacobian matrix of the network itself, and not the norm of the gradient of the cross-entropy loss. Our work differs from the latter work by the fact that we regularize the gradients of the network themselves and not all combinations of their differences, which yields better results  at a lower computational cost, as will be later shown.

We compare the methods mentioned above and adversarial training \cite{Goodfellow15Explaining} to Jacobian regularization on the MNIST, CIFAR-10 and CIFAR-100 datasets, demonstrating the advantage of our strategy in the form of high robustness to the DeepFool \cite{DeepFool}, FGSM \cite{Goodfellow15Explaining}, and JSMA \cite{JSMA} attack methods. Our method surpasses the results of the other strategies on FGSM and DeepFool and achieves competitive performance on JSMA. We also show that using Jacobian regularization combined with adversarial training further improves the robustness results.

This paper is organized as follows. Section~\ref{sec:method} introduces the Jacobian regularization method and related strategies. Section~\ref{sec:theory} shows its connection to some theory of adversarial examples. The relationships drawn in this section suggest that regularizing the Jacobian of deep neural networks can improve their robustness to adversarial examples. In Section~\ref{sec:experiments} we demonstrate empirically the advantages of this approach. Section~\ref{sec:Conclusions} concludes our paper. In the appendices we provide more theoretical insight and additional experimental results.

\section{Jacobian Regularization for Adversarial Robustness}
\label{sec:method}

Adversarial perturbations are essentially small changes in the input data which cause large changes in the network's output. In order to prevent this vulnerability, during the post-processing training phase we penalize large gradients of the classification function with respect to the input data. Thus, we encourage the network's learned function to be more robust to small changes in the input space. This is achieved by adding a regularization term in the form of the Frobenius norm of the network's Jacobian matrix evaluated on the input data. 
The relation between the Frobenius norm and the $\ell_2$ (spectral) norm of the Jacobian matrix has been shown in \cite{Sokolic17Robust}, and lays the justification for using the Frobenius norm of the network's Jacobian as a regularization term. We emphasize that we apply this regularization as additional post-processing training which is done after the regular training has finished.

To describe the Jacobian regularization more formally, we use the following notation. Let us denote the network's input as a $D$-dimensional vector, its output as a $K$-dimensional vector, and let us assume the training dataset $X$ consists of $N$ training examples. We use the index $l=1,...,L$ to specify a certain layer in a network with $L$ layers. $z^{(l)}$ is the output of the $l^{th}$ layer of the network and $z_{k}^{(l)}$ is the output of the $k^{th}$ neuron in this layer.
In addition, let us denote by $\lambda$ the hyper-parameter which controls the weight of our regularization penalty in the loss function. 
The input to the network is
\begin{align}
\label{eq:xi}
x_i \in \mathbb{R}^{D}, ~~~~~ i=1\dots N, ~~~~~ X = \left[
\begin{array}{ccc}
x_1^T \\
\vdots \\
x_N^T \\
\end{array}
\right] \in \mathbb{R}^{N \times D},
\end{align}
and its output is $f(x_i) \in \mathbb{R}^{K}$, where the predicted class $k^*_i$ for an input $x_i$ is $k^*_i = \argmax_{k} f_k(x_i)$, $k = 1,...,K$.

$f(x_i) = \softmax\{z^{(L)}(x_i)\}$ is the network's output after the softmax operation where $z^{(L)}(x_i)$ is the output of the last fully connected layer in the network for the input $x_i$.
The term $\nabla_{x} z^{(L)}(x_i)$ is the Jacobian matrix of layer $L$ evaluated at the point $x_i$, i.e. $J^{(L)}(x_i) = \nabla_{x} z^{(L)}(x_i)$. Correspondingly, $J_k^{(L)}(x_i) = \nabla_{x} z_k^{(L)}(x_i)$ is the $k^{th}$ row in the matrix $J^{(L)}(x_i)$.

A network's Jacobian matrix is given by
\begin{align}
\label{eq:Jacob_matrix}
J(x_i) \triangleq J^{(L)}(x_i) = \left[
\begin{array}{ccc}
\frac{\partial z_1^{(L)}(x_i)}
{\partial x_{(1)}} & \dots & \frac{\partial z_1^{(L)}(x_i)}
{\partial x_{(D)}} \\
\vdots & \ddots & \vdots\\
\frac{\partial z_K^{(L)}(x_i)}
{\partial x_{(1)}} & \dots & \frac{\partial z_K^{(L)}(x_i)}
{\partial x_{(D)}} 
\end{array}
\right] \in \mathbb{R}^{K \times D},
\end{align}
where $x = (x_{(1)} \dots x_{(D)})^T$.
Accordingly, the Jacobian regularization term for an input sample $x_i$ is
\begin{align}
\label{eq:jacob_2}
& ||J(x_i)||_{F}^{2} = \sum_{d=1}^{D} \sum_{k=1}^{K} \left( \frac{\partial}
{\partial x_d} z_k^{(L)}(x_i)\right)^2 = \sum_{k=1}^{K}||\nabla_x z_k^{(L)}(x_i)||_2^2.
\end{align}
Combining the regularization term in \eqref{eq:jacob_2} with a standard cross-entropy loss function on the training data, we get the following loss function for training:
 \begin{align}
 \label{eq:loss_function}
 & Loss = -\sum_{i=1}^{N} \sum_{k=1}^{K} y_{ik}\log{f_{k}(x_i)} + \lambda \sqrt{\sum_{d=1}^{D} \sum_{k=1}^{K} \sum_{i=1}^{N} \left(\frac{\partial}{\partial x_d} z_k^{(L)}(x_i)\right)^2},
 \end{align} 
where $y_{i} \in \mathbb{R}^{K}$ is a one-hot vector representing the correct class of the input $x_i$.

The Input Gradient regularization method from \cite{InputGradients} uses the following regularization term:
\begin{align}
\label{eq:InputGrad}
& \sum_{d=1}^{D} \sum_{i=1}^{N}\left(\frac{\partial}
{\partial x_d} \sum_{k=1}^{K} -y_{ik} \log f_k(x_i)\right)^2.
\end{align}

The Cross-Lipschitz regularization method from \cite{Cross_Lipschitz_regularization} uses the following regularization term:
\begin{align}
\label{eq:CLR}
\sum_{i=1}^{N} \sum_{j,k=1}^{K} ||\nabla_x z_k^{(L)}(x_i) - \nabla_x z_j^{(L)}(x_i)||_2^2.
\end{align}

The adversarial training method \cite{Goodfellow15Explaining} adds perturbed inputs along with their correct labels to the training dataset, so that the network learns the correct labels of perturbed inputs during training. This helps the network to achieve a higher accuracy when it is being fed with new perturbed inputs, meaning the network becomes more robust to adversarial examples.

On the computational complexity aspect, Jacobian regularization introduces an overhead of one additional back-propagation step in every iteration. This step involves the computation of mixed partial derivatives, as the first derivative is w.r.t the input, and the second is w.r.t. the model parameters. However, one should keep in mind that Jacobian regularization is applied as a post-processing phase, and not throughout the entire training, which is computationally beneficial. Moreover, it is also more efficient than the Cross-Lipschitz regularization technique \cite{Cross_Lipschitz_regularization}, which requires the computation of the norm of $\frac{1}{2}K(K-1)$ terms as opposed to our method that only requires the calculation of the norm of $K$ different gradients. This makes Jacobian regularization more scalable for datasets with a large $K$.

\section{Theoretical Justification}
\label{sec:theory}

\subsection{The Jacobian matrix and adversarial perturbations}
In essence, for a network performing a classification task, an adversarial attack (a fooling method) aims at making a change as small as possible, which changes the network's decision. In other words, finding the smallest perturbation that causes the output function to cross a decision boundary to another class, thus making a classification error.
In general, an attack would seek for the closest decision boundary to be reached by an adversarial perturbation in the input space. This makes the attack the least noticeable and the least prone to being discovered \cite{Goodfellow15Explaining}.

To gain some intuition for our proposed defense method, we start with a simple informal explanation on the relationship between adversarial perturbations and the Jacobian matrix of a network.
Let $x$ be a given input data sample; $x_{same}$ a data sample close to $x$ from the same class that was not perturbed by an adversarial attack; and $x_{pert}$  another data sample, which is the result of an adversarial perturbation of $x$ that keeps it close to it but with a different predicted label. Therefore, we have that for the $\ell_2$ distance metric in the input and output of the network
\begin{eqnarray}
\label{eq:perturbed_1}
 \frac{||x_{pert}-x||_2}{||x_{same}-x||_2} \approx 1  & 
 ~~~~\text{and} & 
~~~~ 1<\frac{||z^{(L)}(x_{pert})-z^{(L)}(x)||_2}{||z^{(L)}(x_{same})-z^{(L)}(x)||_2},
\end{eqnarray}
with a high probability. Therefore,
\begin{align}
\label{eq:perturbed_2}
& \frac{||z^{(L)}(x_{same})-z^{(L)}(x)||_2}{||x_{same}-x||_2} <
\frac{||z^{(L)}(x_{pert})-z^{(L)}(x)||_2}{||x_{pert}-x||_2}.
\end{align}
Let $[x,x_{pert}]$ be the $D$-dimensional line in the input space connecting $x$ and $x_{pert}$.
According to the mean value theorem there exists some $x' \in [x,x_{pert}]$ such that
\begin{align}
\label{eq:perturbed_3}
& \frac{||z^{(L)}(x_{pert})-z^{(L)}(x)||^2_2 }{||x_{pert}-x||^2_2} \le \sum_{k=1}^{K} ||\nabla_x z^{(L)}_k(x')||_2^2 = ||J(x')||^2_F.
\end{align}
This suggests that a lower Frobenius norm of the network's Jacobian matrix encourages it to be more robust to small changes in the input space. In other words, the network is encouraged to yield similar outputs for similar inputs.

We empirically examined the average values of the Frobenius norm of the Jacobian matrix of networks trained with various defense methods on the MNIST dataset. The network architecture is described in Section~\ref{sec:experiments}. Table~\ref{table:1} presents these values for both the original inputs and the ones which have been perturbed by DeepFool \cite{DeepFool}. For ``regular'' training with no defense, it can be seen that as predicted, the aforementioned average norm is significantly larger on perturbed inputs. 
Interestingly enough, using adversarial training, which does not regularize the Jacobian matrix directly, decreases the average Frobenius norm of the Jacobian matrix evaluated on perturbed inputs (second row of Table~\ref{table:1}). Yet, when Jacobian regularization is added (with $\lambda = 0.1$), this norm is reduced much more (third and fourth rows of Table~\ref{table:1}). Thus, it is expected to improve the robustness of the network even further. Indeed, this behavior is demonstrated in Section~\ref{sec:experiments}.

\begin{table}[h!]
\centering
\scriptsize
\caption{Average Frobenius norm of the Jacobian matrix at the original data and the data perturbed by DeepFool. DNN is trained on MNIST with various defense methods.}
\begin{tabular}{||c c c||} 
 \hline
 Defense method & $\frac{1}{N} \displaystyle \sum_{i=1}^{N} ||J(x_i)||_F$ & $\frac{1}{N} \displaystyle \sum_{i=1}^{N} ||J(x_{i_{pert}})||_F$ \\ [0.5ex] 
 \hline\hline
 No defense & $0.14$ & $0.1877$ \\ 
 Adversarial Training & $0.141$ & $0.143$ \\
 Jacobian regularization & $0.0315$ & $0.055$ \\
 Jacobian regularization $\&$ Adversarial Training & $0.0301$ & $0.0545$ \\
 \hline
\end{tabular}
\label{table:1}
\end{table}

\subsection{Relation to classification decision boundaries}
\label{relation_to_class_decision_boundaries}

As shown in \cite{DeepFool}, we may locally treat the decision boundaries as hyper-surfaces in the $K$-dimensional output space of the network.
Let us denote $g(x) = w^T x +b = 0$ as a hyper-plane tangent to such a decision boundary hyper-surface in the input space. Using this notion, the following lemma approximates the distance between an input and a perturbed input classified to be at the boundary of a hyper-surface separating between the class of $x$, $k_1$, and another class $k_2$.

\begin{lem}
\label{lem:distance_1_lem}
The first order approximation for the distance between an input $x$, with class $k_1$, and a perturbed input classified to the boundary hyper-surface separating the classes $k_1$ and $k_2$ for an $\ell_2$ distance metric is given by
\begin{align}
\label{eq:distance_2}
& d = {\frac{|z^{(L)}_{k_1}(x)-z^{(L)}_{k_2}(x)|}{||\nabla_{x} z^{(L)}_{k_1}(x) - \nabla_{x} z^{(L)}_{k_2}(x)||_2}}.
\end{align}
\end{lem}

This lemma is given in \cite{DeepFool}. For completeness, we present a short sketch of the proof in Appendix~\ref{Proof Sketch of Lemma 1}. Based on this lemma, the following corollary provides a proxy for the minimal distance that may lead to fooling the network.

\begin{cor}
\label{Corollary2}
Let $k^*$ be the correct class for the input sample $x$. Then the $\ell_2$ norm of the minimal perturbation necessary to fool the classification function is approximated by
\begin{align}
\label{eq:min_distance}
& d^* = \min_{k \ne k^*} {\frac{|z^{(L)}_{k^*}(x)-z^{(L)}_{k}(x)|}{||\nabla_{x} z^{(L)}_{k^*}(x) - \nabla_{x} z^{(L)}_{k}(x)||_2}}.
\end{align}
\end{cor}


To make a direct connection to the Jacobian of the network, we provide the following proposition:
\begin{prop}
\label{prop3}
Let $k^*$ be the correct class for the input sample $x$. Then the first order approximation for the $\ell_2$ norm of the minimal perturbation necessary to fool the classification function is lower bounded by
\begin{align}
\label{lower_bound}
& d^{*} \geq \frac{1}{\sqrt{2}|| J^{(L)}(x)||_F}
\min_{k \neq k^*} |z^{(L)}_{k^*}(x)-z^{(L)}_{k}(x)|.
\end{align}
\end{prop}

The proof of Proposition~\ref{prop3} is given in Appendix~\ref{Proof of Proposition 3}. The term $|z^{(L)}_{k^*}(x)-z^{(L)}_{k}(x)|$ in \eqref{lower_bound} is maximized by the minimization of the cross-entropy term of the loss function, since a DNN aspires to learn the correct output with the largest confidence possible, meaning the largest possible margin in the output space between the correct classification and the other possible classes.
The term $||J^{(L)}(x)||_F$ in the denominator is the Frobenius norm of the Jacobian of the last fully connected layer of the network. It is minimized due to the Jacobian regularization part in the loss function. This is essentially a min-max problem, since we wish to maximize the minimal distance necessary to fool the network, $d^*$. For this reason, applying Jacobian regularization during training increases the minimal distance necessary to fool the DNN, thus providing improved robustness to adversarial perturbations. One should keep in mind that it is important not to deteriorate the network's original test accuracy. This is indeed the case as shown in Section~\ref{sec:experiments}.

An important question is whether the regularization of the Jacobian at earlier layers of the network would yield better robustness to adversarial examples. To this end, we examined imposing the regularization on the $L-1$ and the $L-2$ layers of the network.
Both of these cases generally yielded degraded robustness results compared to imposing the regularization on the last layer of the network. Thus, throughout this work we regularize the Jacobian of the whole network. The theoretical details are given in Appendix~\ref{Jacobian regularization of the network's L-1 layer - Mathematical Analysis} and the corresponding experimental results are given in Appendix~\ref{Jacobian regularization of the network's L-1 and L-2 layers - Experiments}.

\subsection{Relation to decision boundary curvature}
In \cite{Curvature} the authors show the link between a network's robustness to adversarial perturbations and the geometry of its decision boundaries. The authors show that when the decision boundaries are positively curved the network is fooled by small universal perturbations with a higher probability. Here we show that Jacobian regularization promotes the curvature of the decision boundaries to be less positive, thus reducing the probability of the network being fooled by small universal adversarial perturbations.

Let $H_k(x) = \frac{{\partial}^2 z_{k}^{(L)}(x)}{\partial x^2}$ be the Hessian matrix of the network's classification function at the input point $x$ for the class $k$.
As shown in \cite{Curvature}, the decision boundary between two classes $k_1$ and $k_2$ can be locally referred to as the hyper-surface $F_{k_1,k_2}(x) = z^{(L)}_{k_1}(x) - z^{(L)}_{k_2}(x) = 0$. Relying on the work in \cite{ApproxHessian} let us use the approximation $H_k(x) \approx J_{k}(x)^TJ_{k}(x)$ where $J_{k}(x)$ is the $k^{th}$ row in the matrix $J(x)$. The matrix $J_{k}(x)^TJ_{k}(x)$ is a rank one positive semi-definite matrix. Thus, the curvature of the decision boundary $F_{k_1,k_2}(x)$ is given by $x^T(H_{k_1}-H_{k_2})x$, which using the aforementioned approximation, can be approximated by 
\begin{align}
x^T \left(J_{k_1}(x)^TJ_{k_1}(x)-J_{k_2}(x)^TJ_{k_2}(x)\right) x  = \left(J_{k_1}(x)x \right)^2 -\left(J_{k_2}(x)x \right)^2.
\end{align}
Thus, we arrive at the following upper bound for the curvature:
\begin{eqnarray}
\left(J_{k_1}(x)x\right)^2 - \left(J_{k_2}(x)x\right)^2 & \leq & \left(J_{k_1}(x)x\right)^2 + \left(J_{k_2}(x)x\right)^2 \\ 
  &\leq & \sum_{k=1}^K \left(J_k(x)x\right)^2 
  \leq  ||J(x)||^2_F ||x||^2_2,
\end{eqnarray}
where the last inequality stems from the matrix norm inequality. For this reason the regularization of $||J(x)||_F$ promotes a less positive curvature of the decision boundaries in the environment of the input samples. This offers a geometric intuition to the effect of Jacobian regularization on the network's decision boundaries. Discouraging a positive curvature makes a universal adversarial perturbation less likely to fool the classifier.

\section{Experiments}
\label{sec:experiments}
We tested the performance of Jacobian regularization on the MNIST, CIFAR-10 and CIFAR-100 datasets. The results for CIFAR-100, which are generally consistent with the results for MNIST and CIFAR-10, are given in Appendix~\ref{Experimental results for the CIFAR-100 dataset}.
As mentioned before, we use the training with  Jacobian regularization as a post-processing phase to the "regular" training. Using a post-processing training phase is highly beneficial: it has a low additional computational cost as we add the regularization part after the network is already stabilized with a high test accuracy and not throughout the entire training. It also allows taking an existing network and applying the post-processing training phase to it in order to increase its robustness to adversarial examples. We obtained optimal results this way, whereas we found that applying the Jacobian regularization from the beginning of the training yields a lower final test accuracy. 

The improved test accuracy obtained using post-processing training can be explained by the advantage of keeping the original training phase, which allows the network to train solely for the purpose of a high test accuracy. The subsequent post-processing training phase with Jacobian regularization introduces a small change to the already existing good test accuracy, as opposed to the case where the regularization is applied from the beginning that results in a worse test accuracy. Table~\ref{table:9} presents a comparison between post-processing training and "regular" training on MNIST. Similar results are obtained for CIFAR-10 and CIFAR-100.

We examine the performance of our method using three different adversarial attack methods: DeepFool \cite{DeepFool}, FGSM \cite{Goodfellow15Explaining} and JSMA \cite{JSMA}. We also assess the performance of our defense combined with adversarial training, which is shown to be effective in improving the model's robustness. However, this comes at the cost of generating and training on a substantial amount of additional input samples as is the practice in adversarial training. We found that the amount of perturbed inputs in the training mini-batch has an impact on the overall achieved robustness. An evaluation of this matter appears in Appendix~\ref{Influence of the amount of adversarial examples in the training mini-batch}. The results for adversarial training, shown hereafter, are given for the amount of perturbed inputs that yields the optimal results in each test case.
We also compare the results to the Input Gradient regularization technique \cite{InputGradients} and the Cross-Lipschitz regularization technique \cite{Cross_Lipschitz_regularization}.

\begin{table}
\centering
\scriptsize
\caption{Effect of post-processing training vs. "regular" training on MNIST, using different defense methods}
\begin{tabular}{||c c c||} 
 \hline
 Defense method & Test accuracy & $\hat{\rho}_{adv}$ \\ [0.5ex] 
 \hline\hline
 No defense & $99.08\%$ & $20.67$ x ${10^{-2}}$ \\
 Input Gradient regularization, "regular" training & $99.25\%$ & 23.43 x ${10^{-2}}$ \\
 Input Gradient regularization, post-processing training & $99.44\%$ & $24.03$ x ${10^{-2}}$ \\
 Cross-Lipschitz regularization, "regular" training & $98.64\%$ & $29.03$ x ${10^{-2}}$ \\
 Cross-Lipschitz regularization, post-processing training & $98.91\%$ & $29.99$ x ${10^{-2}}$ \\
 Jacobian regularization, "regular" training & $98.35\%$ & $32.89$ x ${10^{-2}}$ \\
 Jacobian regularization, post-processing training & $98.44\%$ & $34.24$ x ${10^{-2}}$ \\
 \hline
\end{tabular}
\label{table:9}
\end{table}

For MNIST we used the network from the official TensorFlow tutorial \cite{abadi2015tensorflow}. The network consists of two convolutional layers, each followed by a max pooling layer. These layers are then followed by two fully connected layers. All these layers use the ReLU activation function, except for the last layer which is followed by a softmax operation. Dropout regularization with 0.5 keep probability is applied to the fully connected layers. The training is done using an Adam optimizer \cite{Kingma15Adam} and a mini-batch size of 500 inputs. With this network we obtained a test accuracy of $99.08\%$. Training with Jacobian regularization was done with a weight of $\lambda = 0.1$, which we found to provide a good balance between the cross-entropy loss and the Jacobian regularization.

For CIFAR-10 we used a convolutional neural network consisting of four concatenated sets, where each set consists of two convolutional layers followed by a max pooling layer followed by dropout with a 0.75 keep probability. After these four sets, two fully connected layers are used. For CIFAR-10, training was done with a RMSProp optimizer \cite{Hinton12RMSprop} and a mini-batch size of 128 inputs. With this network we obtained a test accuracy of $88.79\%$. Training with Jacobian regularization was done with a weight of $\lambda = 0.5$, which we found to provide a good balance between the cross-entropy loss and the Jacobian regularization.

The results of an ablation study regarding the influence of variation in the values of $\lambda$ for MNIST and CIFAR-10 are given in Appendix~\ref{Jacobian regularization - influence of the hyper-parameter lambda}.

\subsection{DeepFool evaluation}

We start by evaluating the performance of our method compared to the others under the DeepFool attack.
The DeepFool attack \cite{DeepFool} uses a first order approximation of the network's decision boundaries as hyper-planes. Using this approximation, the method seeks for the closest decision boundary to be reached by a change in the input. Since the decision boundaries are not actually linear, this process continues iteratively until the perturbed input changes the network's decision. The robustness metric associated with this attack is $\hat{\rho}_{adv} = \frac{1}{N} \sum_{i=1}^{N} \frac{d_i}{||x_i||_2}$, which represents the average proportion between the $\ell_2$ norm of the minimal perturbation necessary to fool the network for an input $x_i$ and the $\ell_2$ norm of $x_i$. This attack is optimized for the $\ell_2$ metric.

Table~\ref{table:2} and Table~\ref{table:3} present the robustness measured by $\hat{\rho}_{adv}$ under a DeepFool attack for MNIST and CIFAR-10 respectively. As the results show, Jacobian regularization provides a much more significant robustness improvement compared to the other methods. Substantially smaller perturbation norms are required to fool networks that use those defense approaches compared to networks that are trained using Jacobian regularization. Moreover, combining it with adversarial training further enhances this difference in the results. 

\begin{table}
\centering
\scriptsize
\caption{Robustness to DeepFool attack for MNIST}
\begin{tabular}{||c c c||} 
 \hline
 Defense method & Test accuracy & $\hat{\rho}_{adv}$ \\ [0.5ex] 
 \hline\hline
No defense & $99.08\%$ & $20.67$ x ${10^{-2}}$ \\ 
Adversarial Training & $99.03\%$ & $22.38$ x ${10^{-2}}$ \\
Input Gradient regularization & $99.25\%$ & 23.43 x ${10^{-2}}$ \\
Input Gradient regularization $\&$ Adversarial Training & $98.88\%$ & 23.49 x ${10^{-2}}$ \\
Cross-Lipschitz regularization & $98.64\%$ & $$29.03 x ${10^{-2}}$ \\
Cross-Lipschitz regularization $\&$ Adversarial Training & $98.73\%$ & $$32.38 x ${10^{-2}}$ \\
Jacobian regularization & $98.44\%$ & $34.24$ x ${10^{-2}}$ \\
Jacobian regularization $\&$ Adversarial Training & $98\%$ & $36.29$ x ${10^{-2}}$ \\
 \hline
\end{tabular}
\label{table:2}
\end{table}

\begin{table}
\centering
\scriptsize
\caption{Robustness to DeepFool attack for CIFAR-10}
\begin{tabular}{||c c c||} 
\hline
Defense method & Test accuracy & $\hat{\rho}_{adv}$ \\ [0.5ex] 
\hline\hline
No defense & $88.79\%$ & $1.21$ x ${10^{-2}}$ \\
Adversarial Training & $88.88\%$ & $1.23$ x ${10^{-2}}$ \\
Input Gradient regularization & $88.56\%$ & $1.43$ x ${10^{-2}}$ \\
Input Gradient regularization $\&$ Adversarial Training & $88.49\%$ & $2.17$ x ${10^{-2}}$ \\  
Cross-Lipschitz regularization & $88.91\%$ & $2.08$ x ${10^{-2}}$ \\
Cross-Lipschitz regularization $\&$ Adversarial Training & $88.49\%$ & $4.04$ x ${10^{-2}}$ \\  
Jacobian regularization & $89.16\%$ & $3.42$ x ${10^{-2}}$ \\
 Jacobian regularization $\&$ Adversarial Training & $88.49\%$ & $6.03$ x ${10^{-2}}$ \\
 \hline
\end{tabular}
\label{table:3}
\end{table}


Notice that neither of the examined defense methods change the test accuracy significantly. For MNIST, the Jacobian and Cross-Lipschitz regularizations and adversarial training cause a small accuracy decrease, whereas the Input Gradient regularization technique improves the accuracy. Conversely, for CIFAR-10, the Jacobian and Cross-Lipschitz regularizations and adversarial training yield a better accuracy, whereas the Input Gradient regularization reduces the accuracy.


\subsection{FGSM evaluation}

The FGSM (Fast Gradient Sign Method) attack \cite{Goodfellow15Explaining} was designed to rapidly create adversarial examples that could fool the network. The method changes the network's input according to:
\begin{align}
x_{pert} = x + \epsilon \cdot sign \left( \nabla_x Loss(x) \right),
\end{align}
where $\epsilon$ represents the magnitude of the attack. This attack is optimized for the $\ell_{\infty}$ metric.

\begin{figure}
  \centering
  \begin{subfigure}{0.5\linewidth}%
  \centering
  \includegraphics[scale=0.16]{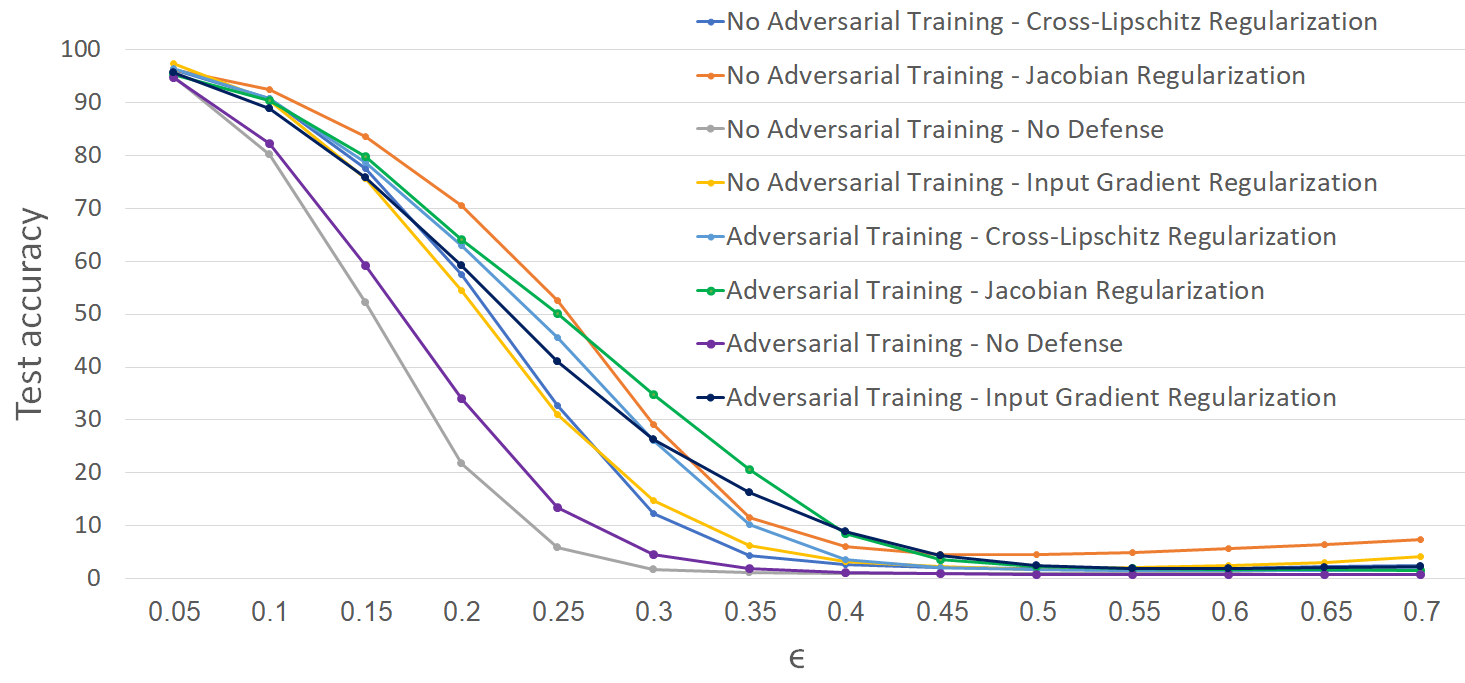}
  \caption{MNIST}
  \label{fig:MNIST_FGSM}%
  \end{subfigure}%
  \hfill
  \begin{subfigure}{0.5\linewidth}%
  \centering
  \includegraphics[scale=0.17]{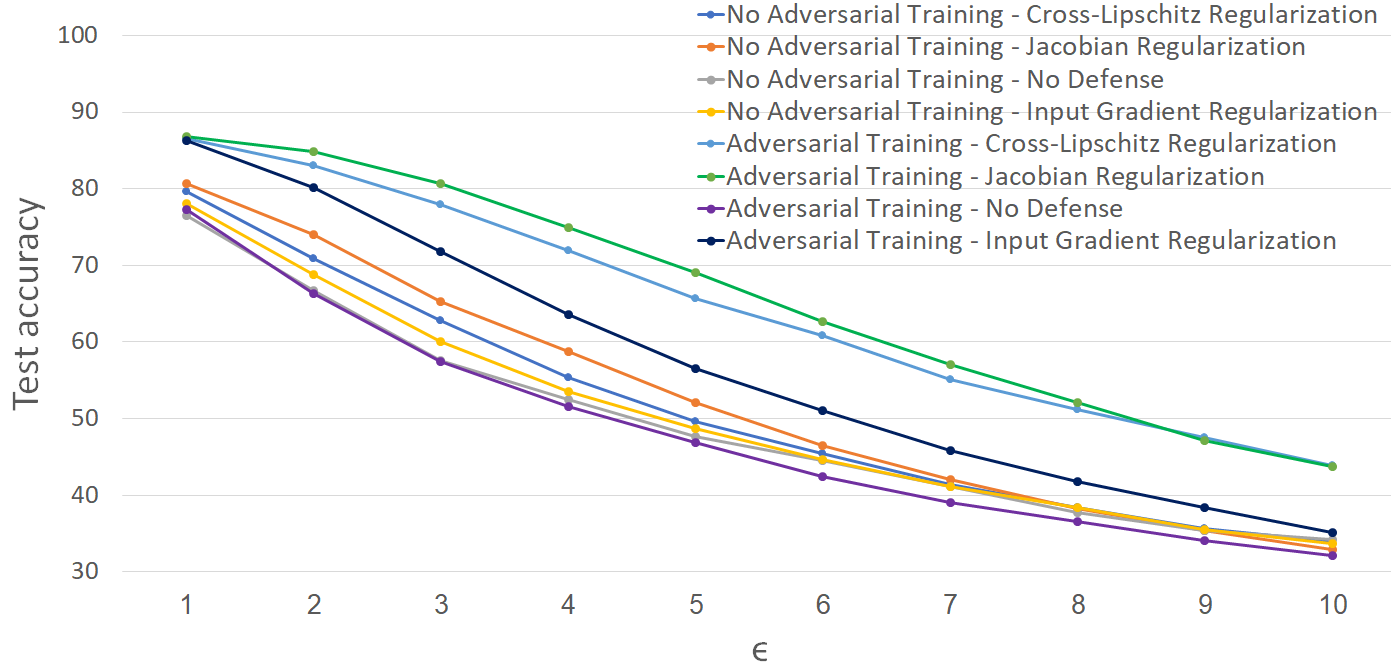}
  \caption{CIFAR-10}
  \label{fig:CIFAR-10_FGSM}%
  \end{subfigure}%
  \caption{Test accuracy for FGSM attack on MNIST (left) and CIFAR-10 (right) for different values of $\epsilon$}
  \label{fig:FGSM}%
\end{figure}

We examined the discussed defense methods' test accuracy under the FGSM attack (test accuracy on the perturbed dataset) for different values of $\epsilon$.
Fig.~\ref{fig:FGSM} presents the results comparing Jacobian regularization to adversarial training, Input Gradient regularization and Cross-Lipschitz regularization. In all cases, the minimal test accuracy on the original test set using Jacobian regularization is $98\%$ for MNIST and $88.49\%$ for CIFAR-10.

Similarly to the results under the DeepFool attack, the results under the FGSM attack show that the test accuracy with the Jacobian regularization defense is higher than with the Input Gradient and Cross-Lipschitz regularizations or with adversarial training. Moreover, if adversarial training is combined with Jacobian regularization, its advantage over using the other techniques is even more distinct. This leads to the conclusion that the Jacobian regularization method yields a more robust network to the FGSM attack.


\subsection{JSMA evaluation}

The JSMA (Jacobian-based Saliency Map Attack) \cite{JSMA} attack relies on the computation of a Saliency Map, which outlines the impact of every input pixel on the classification decision. The method picks at every iteration the most influential pixel to be changed such that the likelihood of the target class is increased. We leave the mathematical details to the original paper. Similarly to FGSM, $\epsilon$ represents the magnitude of the attack. The attack is repeated iteratively, and is optimized for the $\ell_0$ metric.

\begin{figure}
  \centering
  \begin{subfigure}{0.5\linewidth}%
  \centering
  \includegraphics[scale=0.17]{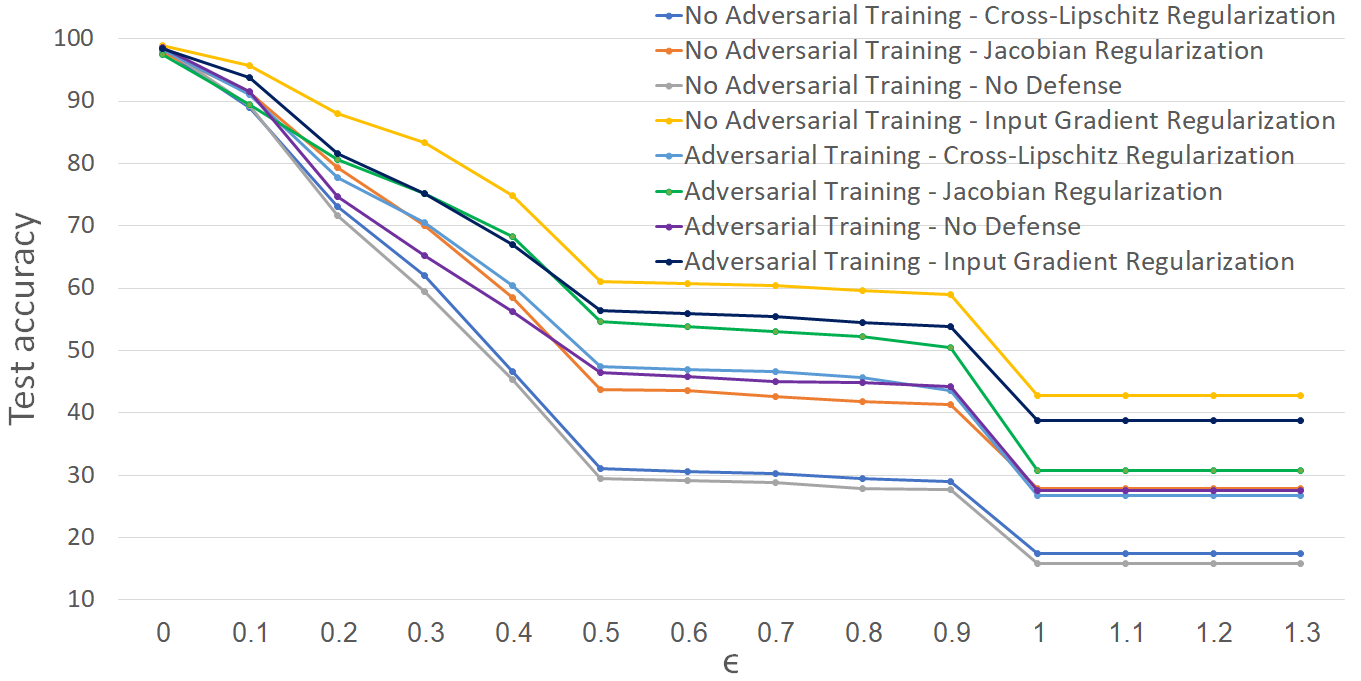}
  \caption{MNIST}
  \label{fig:MNIST_JSMA}%
  \end{subfigure}%
  \hfill
  \begin{subfigure}{0.5\linewidth}%
  \centering
  \includegraphics[scale=0.18]{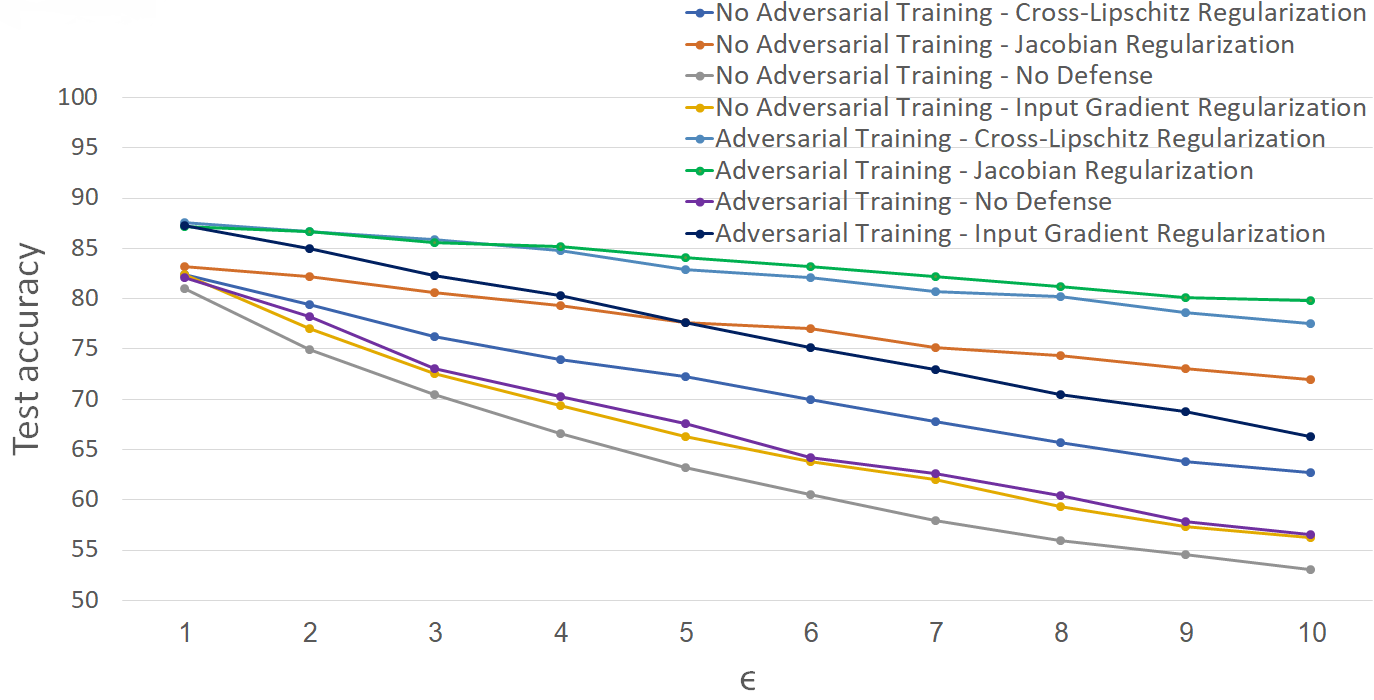}
  \caption{CIFAR-10}
  \label{fig:CIFAR-10_JSMA}%
  \end{subfigure}%
  \caption{Test accuracy for JSMA (1 pixel) attack on MNIST with 80 epochs (left) and CIFAR-10 with 200 epochs (right) for different values of $\epsilon$}
  \label{fig:JSMA}%
\end{figure}

We examined the defense methods' test accuracy under the JSMA attack (test accuracy on the perturbed dataset) for different values of $\epsilon$. Fig.~\ref{fig:JSMA} presents the results for the MNIST and CIFAR-10 datasets. The parameters of the JSMA attack are 80 epochs, 1 pixel attack for the former and 200 epochs, 1 pixel attack, for the latter.
In all cases, the minimal test accuracy on the original test set using Jacobian regularization is $98\%$ for MNIST and $88.49\%$ for CIFAR-10.

Our method achieves superior results compared to the other three methods on CIFAR-10. On the other hand, on MNIST we obtain an inferior performance compared to the Input Gradient regularization method, though we obtain a better performance compared to Cross-Lipschitz regularization. Thus, we conclude that our defense method is effective under the JSMA attack in some cases and presents competitive performance overall. 
We believe that the reason behind the failure of our method in the MNIST case can be explained by our theoretical analysis. In the formulation of the Jacobian regularization (based on the Frobenius norm of the Jacobian matrix), the metric that is being minimized is the $\ell_2$ norm. Yet, in the JSMA attack, the metric that is being targeted by the perturbation is the $\ell_0$ pseudo-norm as only one pixel is being changed in every epoch. We provide more details on this issue in Appendix~\ref{Jacobian regularization and the JSMA attack}. 


\section{Discussion and Conclusions}
\label{sec:Conclusions}
This paper introduced the Jacobian regularization method for improving DNNs' robustness to adversarial examples. We provided a theoretical foundation for its usage and showed that it yields a high degree of robustness, whilst preserving the network's test accuracy. We demonstrated its improvement in reducing the vulnerability to various adversarial attacks (DeepFool, FGSM and JSMA) on the MNIST, CIFAR-10 and CIFAR-100 datasets.
Under all three examined attack methods Jacobian regularization exhibits a large improvement in the network's robustness to adversarial examples, while only slightly changing the network's performance on the original test set.
Moreover, in general, Jacobian regularization without adversarial training is better than adversarial training without Jacobian regularization, whereas the combination of the two defense methods provides even better results.
Compared to the Input Gradient regularization, our proposed approach achieves superior performance under two out of the three attacks and competitive ones on the third (JSMA).
Compared to the Cross-Lipschitz regularization, our proposed approach achieves superior performance under all of the three examined attacks.

We believe that our approach, with its theoretical justification, may open the door to other novel strategies for defense against adversarial attacks. 

In the current form of regularization of the Jacobian, its norm is evaluated at the input samples. We empirically deduced that the optimal results are obtained by applying the Jacobian regularization on the original input samples, which is also more efficient computationally, and not on perturbed input samples or on points in the input space for which the Frobenius norm of the Jacobian matrix is maximal. A future work may analyze the reasons for that. 

Notice that in the Frobenius norm, all the rows of the Jacobian matrix are penalized equally. Another possible future research direction is providing a different weight for each row. This may be achieved by either using a weighted version of the Frobenius norm or by replacing it with other norms such as the spectral one. Note, though, that the latter option is more computationally demanding compared to our proposed approach.

\noindent {\bf Acknowledgment.} This work is partially supported by the ERC-StG SPADE grant and the MAGNET MDM grant.

\appendix


\section*{Appendix}

\section{Proof Sketch of Lemma 1}
\label{Proof Sketch of Lemma 1}

\noindent \textbf{Lemma 1}
\label{Lemma1_reiterated}
\emph{The first order approximation for the distance between an input $x$, with class $k_1$, and a perturbed input classified to the boundary hyper-surface separating the classes $k_1$ and $k_2$ for an $\ell_2$ distance metric is given by}
\begin{align}
\label{eq:distance_2_reiterated}
& d = {\frac{|z^{(L)}_{k_1}(x)-z^{(L)}_{k_2}(x)|}{||\nabla_{x} z^{(L)}_{k_1}(x) - \nabla_{x} z^{(L)}_{k_2}(x)||_2}}.
\end{align}

\noindent \emph{Proof Sketch.}
Let $g(x)=w^T x +b=0$ be a hyper-plane tangent to a decision boundary hyper-surface separating between two classes $k_1$ and $k_2$ in the input space. Let the point of tangency be $x_0$ such that $g(x_0) = w^Tx_0+b=0$. The distance between a point $x$ and the hyper-plane $g(x)$ is given by
\begin{align}
\label{eq:distance_1_reiterated}
& d = {\frac{|w^T(x-x_0)|}{||w||_2}} = \frac{|w^Tx+b|}{||w||_2} = \frac{|g(x)|}{||w||_2} = \frac{|g(x)|}{||\nabla_{x} g(x)||_2}.
\end{align}
Since $g(x_0)$ is on the boundary hyper-surface between the classes $k_1$ and $k_2$ it holds that
\begin{align}
\label{eq:hyperplane_reiterated}
& g(x_0) = z^{(L)}_{k_1}(x_0) - z^{(L)}_{k_2}(x_0) = 0.
\end{align}
For a point $x$, which is in the environment of $x_0$
\begin{align}
\label{eq:hyperplane_2_reiterated}
& g(x) \approx z^{(L)}_{k_1}(x) - z^{(L)}_{k_2}(x).
\end{align}
From \eqref{eq:distance_1_reiterated} and \eqref{eq:hyperplane_2_reiterated} it follows that the first order approximation of the distance (for the $\ell_2$ metric) between an input $x$, with class $k_1$, and a perturbed input classified to the boundary hyper-surface separating the classes $k_1$ and $k_2$ is given by \eqref{eq:distance_2_reiterated}. \qed

\section{Proof of Proposition 3}
\label{Proof of Proposition 3}

We reiterate Corollary 2 and Proposition 3 before proving the latter.

\singlespacing
\noindent \textbf{Corollary 2}
\label{Corollary2_reiterated}
\emph{Let $k^*$ be the correct class for the input sample $x$. Then the $\ell_2$ norm of the minimal perturbation necessary to fool the classification function is approximated by}
\begin{align}
& d^* = \min_{k \ne k^*} {\frac{|z^{(L)}_{k^*}(x)-z^{(L)}_{k}(x)|}{||\nabla_{x} z^{(L)}_{k^*}(x) - \nabla_{x} z^{(L)}_{k}(x)||_2}}.
\end{align}


\noindent Proposition~\ref{prop3} makes a direct connection to the Jacobian of the network.

\singlespacing
\noindent \textbf{Proposition 3}
\label{prop3_reiterated}
\emph{Let $k^*$ be the correct class for the input sample $x$. Then the first order approximation for the $\ell_2$ norm of the minimal perturbation necessary to fool the classification function is lower bounded by}
\begin{align}
\label{lower_bound_prop_3}
& d^{*} \geq \frac{1}{\sqrt{2}|| J^{(L)}(x)||_F}
\min_{k \neq k^*} |z^{(L)}_{k^*}(x)-z^{(L)}_{k}(x)|.
\end{align}

\noindent \emph{Proof.}
Relying on Corollary~\ref{Corollary2} we get that
\begin{eqnarray}
d^{*^2}  & = & \min_{k \ne k^*} {\frac{|z^{(L)}_{k^*}(x)-z^{(L)}_{k}(x)|^2}{||\nabla_{x} z^{(L)}_{k^*}(x) - \nabla_{x} z^{(L)}_{k}(x)||^2_2}}
= \min_{k \ne k^*} \frac{|z^{(L)}_{k^*}(x)-z^{(L)}_{k}(x)|^2}
{||J^{(L)}_{k^*}(x) - J^{(L)}_{k}(x)||^2_2}
\\
& = & \min_{k \ne k^*} \frac{|z^{(L)}_{k^*}(x)-z^{(L)}_{k}(x)|^2}
{||J^{(L)}_{k^*}(x)||^2_2 -2J^{(L)}_{k^*}(x)J^{(L)}_{k}(x)^{T} + ||J^{(L)}_{k}(x)||^2_2}
\\
& \geq & \min_{k \ne k^*} \frac{|z^{(L)}_{k^*}(x)-z^{(L)}_{k}(x)|^2}
{2\left(||J^{(L)}_{k^*}(x)||^2_2 + ||J^{(L)}_{k}(x)||^2_2\right)}.
\end{eqnarray}
Since $\sum_{k=1}^K ||J^{(L)}_{k}(x)||^2_2 = ||J^{(L)}(x)||^2_F$ we get that
\begin{align}
& d^{*^2} \geq \frac{1}{2||J^{(L)}(x)||^2_F} \min_{k \ne k^*} |z^{(L)}_{k^*}(x)-z^{(L)}_{k}(x)|^2,
\end{align}
and accordingly,
\begin{align}
& d^{*} \geq \frac{1}{\sqrt{2}||J^{(L)}(x)||_F} \min_{k \ne k^*} |z^{(L)}_{k^*}(x)-z^{(L)}_{k}(x)|
\end{align}
as stipulated. \qed

\section{Jacobian regularization of the network's $L-1$ layer - Mathematical Analysis}
\label{Jacobian regularization of the network's L-1 layer - Mathematical Analysis}

To provide a bound for the $L-1$ layer of the network, we rely on the work in \cite{Orthonormal}, which shows that fixating the weight matrix of the last fully connected layer in a deep neural network causes little to no loss of accuracy while allowing memory and computational benefits. Assuming that this layer corresponds to a weight matrix with $K$ orthonormal columns, it is possible to take another path in the proof of Proposition~\ref{prop3} and obtain a bound as a function of the Jacobian of the $L-1$ layer of the network. This bound is exactly as \eqref{lower_bound_prop_3}, but with $J^{(L-1)}(x)$ instead of $J^{(L)}(x)$, as formulated in Proposition~\ref{prop4} hereafter. 
Note that a trivial application of the multiplicative matrix norm inequality on \eqref{lower_bound_prop_3} leads to a bound with a factor of $\sqrt{K}$ in the denominator since
\begin{eqnarray}
\label{eq:prop3_prop4_1}
 ||J^{(L)}(x)||_F = ||W^{(L)^T} J^{(L-1)}(x) ||_F
&\leq & ||W^{(L)^T}||_F ||J^{(L-1)}(x) ||_F \\ &=& \sqrt{K} ||J^{(L-1)}(x) ||_F,
\end{eqnarray}
where the last equality stems from the fact that $||W^{(L)}||^2_F = K$ for an orthonormal $W^{(L)}$ with $K$ columns.



Given that we have two similar bounds, one depending on the Jacobian of the whole network and one on the Jacobian of the $L-1$ layer, it is important to ask which of them should we regularize to render better robustness to the network. 
The experimental results for the regularization of the Jacobian of the $L-1$ layer of the network, with and without a fixed $W^{(L)}$ with orthonormal columns, are given in Appendix~\ref{Jacobian regularization of the network's L-1 and L-2 layers - Experiments}.
In this section and the following we use the following additional notation:
\begin{itemize}
\item
The network's $L-1$ layer is a fully connected layer consisting of $M$ neurons. We use the index $m = 1,..., M$ for a specific neuron in this layer.
\item
$J^{(L-1)}(x) = \nabla_{x}z^{(L-1)}(x) =
\left[
\begin{array}{ccc}
\nabla_{x}z^{(L-1)}_1(x) \\
\vdots \\
\nabla_{x} z^{(L-1)}_M(x) \\
\end{array}
\right] \in \mathbb{R}^{M \times D}$ is the Jacobian matrix of layer $L-1$ of the network with respect to its input $x$, $J_m^{(L-1)}(x) = \nabla_{x}z^{(L-1)}_m(x)$ is the $m^{th}$ row in this matrix.
\item
We assume that no activation or other non-linear function is applied before the softmax. Thus, the relationship between the $L-1$ layer and the last layer of the network is as follows:
$ z^{(L)} = W^{(L)^{T}}z^{(L-1)}+b^{(L)}$, $W^{(L)} \in \mathbb{R}^{M \times K}$, $b^{(L)} \in \mathbb{R}^{K \times 1}$, where $W_k^{(L)}$ is the $k^{th}$ column in the matrix and $k=1,...,K$.
\end{itemize}

We introduce the following proposition to lay the theoretical foundation for the regularization of the Jacobian matrix of the $L-1$ layer of the network.
\begin{prop}
\label{prop4}
Let $k^*$ be the correct class for the input sample $x$ and let $W^{(L)}$, the weight matrix of the last fully connected layer in the network,  have $K$ orthonormal columns. Then, the first order approximation for $d^{*}$, the $\ell_2$-norm of the minimal perturbation necessary to fool the classification function, is lower bounded as
\begin{align}
\label{eq:penultimate_bound}
& d^{*} \geq \frac{1}{{\sqrt{2}|| J^{(L-1)}(x)||_F}}
\min_{k \neq k^*} |z^{(L)}_{k^*}(x)-z^{(L)}_{k}(x)|.
\end{align}
\end{prop}

\noindent \emph{Proof.}
Since the weight matrix $W^{(L)}$ has $K$ orthonormal columns, then for any $k_1 \ne k_2$ it holds that
$W^{(L)^{T}}_{k_1} W^{(L)}_{k_2} = 0$ and for any $k$ it holds that $W^{(L)^{T}}_{k} W^{(L)}_{k} = 1$.

We remind the reader that according to Lemma~\ref{lem:distance_1_lem}, the first order approximation for the $\ell_2$ distance between an input $x$, with class $k_1$, and a perturbed input classified to the boundary hyper-surface separating the classes $k_1$ and $k_2$ is given by
\begin{eqnarray}
d & = & {\frac{|z_{k_1}^{(L)}(x)-z_{k_2}^{(L)}(x)|}{||\nabla_{x} z_{k_1}^{(L)}(x) - \nabla_{x} z_{k_2}^{(L)}(x)||_2}}.
\end{eqnarray}
Developing this equality further using the chain rule, we have
\begin{eqnarray}
d & = &  \frac{|z^{(L)}_{k_{1}}(x)-z^{(L)}_{k_{2}}(x)|}
{|| \sum_{m=1}^{M} \frac{\partial z_{k_{1}}^{(L)}(x)}{\partial z_{m}^{(L-1)}}
\frac{\partial z_{m}^{(L-1)}}{\partial x}
- \sum_{m=1}^{M} \frac{\partial z_{k_{2}}^{(L)}(x)}{\partial z_{m}^{(L-1)}} \frac{\partial z_{m}^{(L-1)}}{\partial x}||_2} 
\\
& = & \frac{|z^{(L)}_{k_{1}}(x)-z^{(L)}_{k_{2}}(x)|}
{|| \sum_{m=1}^{M} \left( W_{m,k_1}^{(L)} - W_{m,k_2}^{(L)} \right) \nabla_{x} z_{m}^{(L-1)}(x) ||_2}
\\
\label{eq:d_bound_orthogonal_step1}
& = & \frac{|z^{(L)}_{k_{1}}(x)-z^{(L)}_{k_{2}}(x)|}
{|| ( W_{k_1}^{(L)^T} - W_{k_2}^{(L)^T} ) J^{(L-1)}(x) ||_2} \geq
\frac{|z^{(L)}_{k_{1}}(x)-z^{(L)}_{k_{2}}(x)|}
{|| J^{(L-1)}(x)||_F ||W_{k1}^{(L)} - W_{k_{2}}^{(L)} ||_2},
\end{eqnarray}
where the last inequality stems from the multiplicative matrix norm inequality.
Since the columns of $W^{(L)}$ are orthonormal it holds that
\begin{align}
\label{eq:Wk1_Wk2_orthogonal}
& ||W_{k_{1}}^{(L)} - W_{k_{2}}^{(L)} ||^{2}_{2} = 
||W_{k_{1}}^{(L)} ||^{2}_{2} + ||W_{k_{2}}^{(L)} ||^{2}_{2} = 2.
\end{align}
Plugging \eqref{eq:Wk1_Wk2_orthogonal} in \eqref{eq:d_bound_orthogonal_step1} leads to
\begin{align}
& d  \geq
\frac{|z^{(L)}_{k_{1}}(x)-z^{(L)}_{k_{2}}(x)|}
{\sqrt{2}|| J^{(L-1)}(x)||_F}.
\end{align}

Let $k^*$ be the correct class for the input sample $x$. Then, the first order approximation for the $\ell_2$-norm of the minimal perturbation necessary to fool the network is exactly as lower bounded in \eqref{eq:penultimate_bound}. \qed

\section{Jacobian regularization of the network's $L-1$ and $L-2$ layers - Experiments}
\label{Jacobian regularization of the network's L-1 and L-2 layers - Experiments}

In this section we show the empirical results of a regularization based on the Frobenius norm of the Jacobian matrix of the $L-1$ and $L-2$ layers of the network ($J^{(L-1)}(x)$ and $J^{(L-2)}(x)$ respectively).

Since the $L-1$ layer typically consists of substantially more neurons than the last layer, i.e. $M \gg K$, the evaluation of the Jacobian matrix of the $L-1$ layer is much more computationally demanding. For example, in our network for MNIST classification, $M=1024 \gg K = 10$. 
Accordingly, when regularizing the Jacobian matrix of the $L-1$ layer we reduced the size of the training mini-batch to 50 inputs per mini-batch due to computational constraints.

This increase in the computational overhead required for the regularization of the Jacobian of the network's $L-1$ layer is a significant factor in our decision to prefer the regularization of the last layer of the Jacobian matrix. Our choice of the last layer is further supported by the fact that it also leads to superior results under most attack methods  compared to the regularization of the $L-1$ layer as we show hereafter.

We examine two cases: a case in which the weight matrix $W^{(L)}$ is fixed with $K$ orthonormal columns (i.e. not updated during training), and a case in which no restriction is imposed on $W^{(L)}$ and it is updated during training.
For comparison, Table~\ref{table:4} (same as Table~\ref{table:2}) presents the results achieved for MNIST under the DeepFool attack with the regularization of the Frobenius norm of the Jacobian of the last layer of the network.

\begin{table}
\centering
\scriptsize
\caption{Robustness to DeepFool attack for MNIST, regularization based on the Jacobian of the last layer}
\begin{tabular}{||c c c||} 
 \hline
 Defense method & Test accuracy & $\hat{\rho}_{adv}$ \\ [0.5ex] 
 \hline\hline
 No defense & $99.08\%$ & $20.67$ x ${10^{-2}}$ \\ 
 Adversarial Training & $99.03\%$ & $22.38$ x ${10^{-2}}$ \\
 Jacobian regularization & $98.44\%$ & $34.24$ x ${10^{-2}}$ \\
    Jacobian regularization $\&$ Adversarial Training & $98\%$ & $36.29$ x ${10^{-2}}$ \\
 \hline
\end{tabular}
\label{table:4}
\end{table}

\begin{table}
\centering
\scriptsize
\caption{Robustness to DeepFool attack for MNIST -- regularization of $J^{(L-1)}(x)$, with a fixed $W^{(L)}$ with $K$ orthonormal columns}
\begin{tabular}{||c c c||} 
 \hline
 Defense method & Test accuracy & $\hat{\rho}_{adv}$ \\ [0.5ex] 
 \hline\hline
 No defense & $98.23\%$ & $20.18$ x ${10^{-2}}$ \\ 
 Adversarial Training & $98.06\%$ & $24.20$ x ${10^{-2}}$ \\
 Jacobian regularization & $97.10\%$ & $29.61$ x ${10^{-2}}$ \\
    Jacobian regularization $\&$ Adversarial Training & $96.67\%$ & $31.06$ x ${10^{-2}}$ \\
 \hline
\end{tabular}
\label{table:5}
\end{table}

\begin{table}
\centering
\scriptsize
\caption{Robustness to DeepFool attack for MNIST -- regularization of $J^{(L-1)}(x)$, $W^{(L)}$ updated during training}
\begin{tabular}{||c c c||} 
 \hline
 Defense method & Test accuracy & $\hat{\rho}_{adv}$ \\ [0.5ex] 
 \hline\hline
 No defense & $99.08\%$ & $20.67$ x ${10^{-2}}$ \\ 
 Adversarial Training & $99.03\%$ & $22.38$ x ${10^{-2}}$ \\
 Jacobian regularization & $98.75\%$ & $28.16$ x ${10^{-2}}$ \\
 Jacobian regularization $\&$ Adversarial Training & $98.54\%$ & $32.02$ x ${10^{-2}}$ \\
 \hline
\end{tabular}
\label{table:6}
\end{table}

Tables~\ref{table:5} and \ref{table:6} present the results for MNIST under the DeepFool attack with a regularization based on the Frobenius norm of the Jacobian of the $L-1$ layer of the network $J^{(L-1)}(x)$.
Table~\ref{table:5} considers the case where the last layer of the network has a fixed weight matrix $W^{(L)}$ with $K$ orthonormal columns and Table~\ref{table:6} demonstrates the scenario where the weight matrix $W^{(L)}$ is updated during training.  

Note that regularizing the Jacobian of the last layer of the network is significantly better compared to the cases where the $L-1$ layer is regularized. In addition, it is interesting to remark that when $W^{(L)}$ is updated during training, the test accuracy on the original dataset is higher compared to the case in which $W^{(L)}$ is fixed. Yet, the robustness results under the DeepFool attack are similar in both cases.

\begin{figure*}
  \centering
  \begin{subfigure}[b]{0.5\linewidth}%
  \centering
  \includegraphics[scale=0.16]{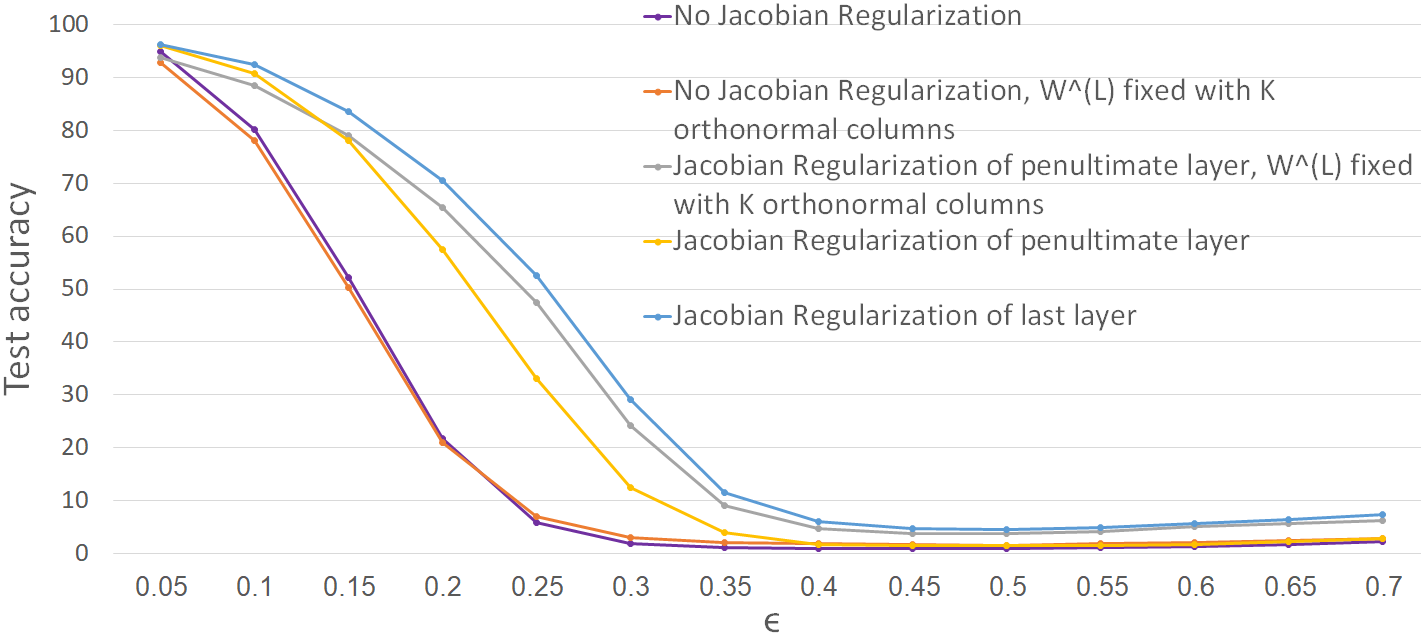}
  \caption{FGSM attack}
  \label{fig:MNIST_FGSM_J_L-1}
  \end{subfigure}%
  \hfill
  \begin{subfigure}[b]{0.5\linewidth}%
  \centering
  \includegraphics[scale=0.155]{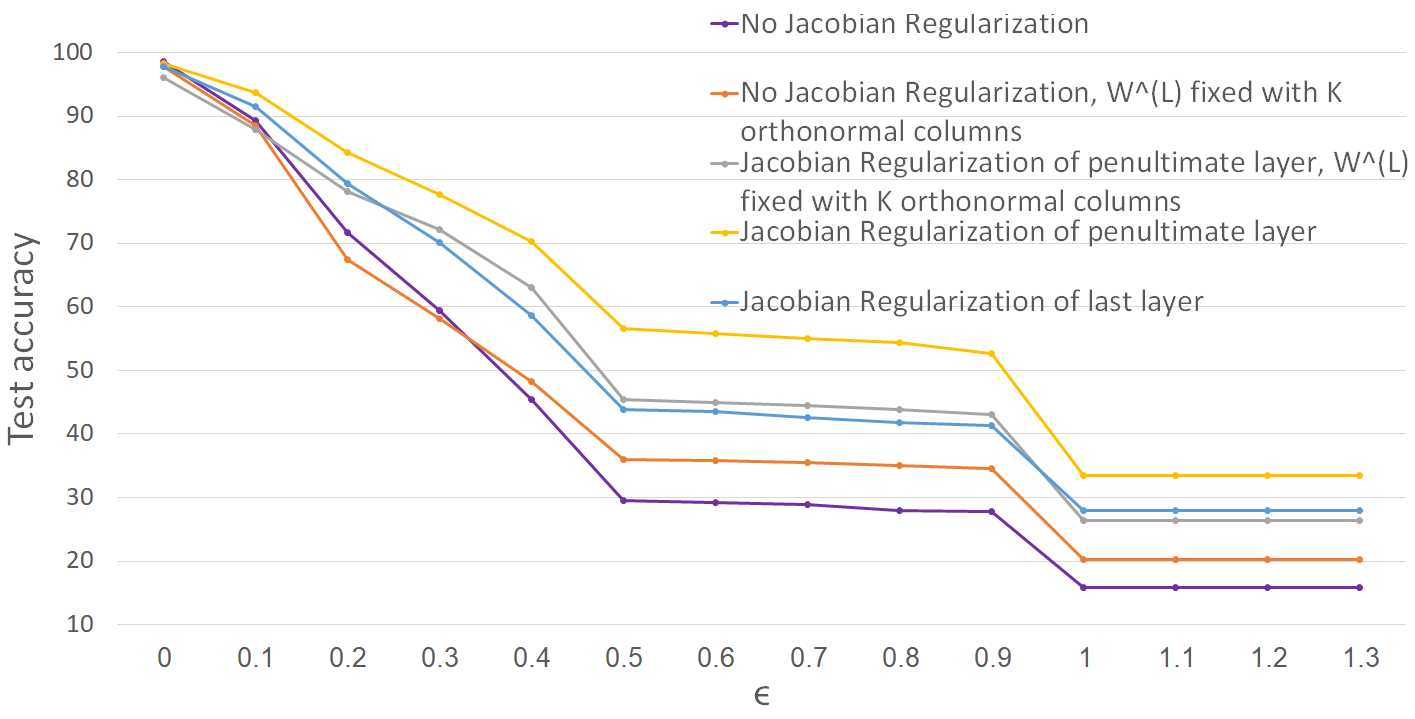}
  \caption{JSMA attack}
  \label{fig:MNIST_JSMA_J_L-1}
  \end{subfigure}%
  \caption{MNIST test accuracy under FGSM (left) and JSMA (right) attacks for different values of $\epsilon$ for Jacobian regularization of layer $L$; Jacobian regularization of layer $L-1$ with $W^{(L)}$ fixed with $K$ orthonormal columns; and Jacobian regularization of layer $L-1$ with $W^{(L)}$ updated during training}
\label{fig:MNIST_ATTACKS_J_L-1}
\end{figure*}

Comparisons of the test accuracies under the FGSM and JSMA attack methods are presented in Fig.~\ref{fig:MNIST_ATTACKS_J_L-1}. The JSMA attack was performed as a 1 pixel attack with 80 epochs.

Note that under the JSMA attack, the regularization of the network's $L-1$ layer yields better results, whereas under the FGSM attack the regularization of the last layer of the network yields better results.
Maintaining the weight matrix $W^{(L)}$ constant with $K$ orthonormal columns generally harms the robustness results in the former case and improves the robustness results in the latter case.

For completeness, we also examined the case where Jacobian regularization is applied to the $L-2$ layer of the network, i.e. regularizing the Frobenius norm of $J^{(L-2)}(x)$, both with $W^{(L)}$ and $W^{(L-1)}$ updated during training and with $W^{(L)}$ and $W^{(L-1)}$ fixed with $K$ and $M$ orthonormal columns respectively. This case is even more computationally demanding, e.g. in the case of MNIST our network's $L-2$ layer consists of 3136 neurons. The empirical results under the DeepFool attack are given in Table~\ref{table:12} and Table~\ref{table:13}. 

\begin{table}
\centering
\scriptsize
\caption{Robustness to DeepFool attack for MNIST -- regularization of $J^{(L-2)}(x)$, with fixed $W^{(L)}$ and $W^{(L-1)}$ with $K$ and $M$ orthonormal columns respectively}
\begin{tabular}{||c c c||} 
\hline
Defense method & Test accuracy & $\hat{\rho}_{adv}$ \\ [0.5ex] 
\hline\hline
No defense & $95.45\%$ & $17.60$ x ${10^{-2}}$ \\ 
Adversarial Training & $95.26\%$ & $20.83$ x ${10^{-2}}$ \\
Jacobian regularization & $94.86\%$ & $22.52$ x ${10^{-2}}$ \\
Jacobian regularization $\&$ Adversarial Training & $94.50\%$ & $24.21$ x ${10^{-2}}$ \\
\hline
\end{tabular}
\label{table:12}
\end{table}

\begin{table}
\centering
\scriptsize
\caption{Robustness to DeepFool attack for MNIST -- regularization of $J^{(L-2)}(x)$, $W^{(L)}$ and $W^{(L-1)}$ updated during training}
\begin{tabular}{||c c c||} 
\hline
Defense method & Test accuracy & $\hat{\rho}_{adv}$ \\ [0.5ex] 
\hline\hline
No defense & $99.08\%$ & $20.67$ x ${10^{-2}}$ \\ 
Adversarial Training & $99.03\%$ & $22.38$ x ${10^{-2}}$ \\
Jacobian regularization & $98.51\%$ & $23.19$ x ${10^{-2}}$ \\
Jacobian regularization $\&$ Adversarial Training & $98.14\%$ & $25.41$ x ${10^{-2}}$ \\
\hline
\end{tabular}
\label{table:13}
\end{table}

As can be seen in the results, the obtained robustness is significantly lower when Jacobian regularization is applied to the $L-2$ layer of the network, compared to the $L-1$ layer and the last layer. The results show that Jacobian regularization becomes less effective when it is applied to earlier layers of the network. In the case where $W^{(L)}$ and $W^{(L-1)}$ are updated during training, this can be explained by the network's increased ability to compensate for this regularization using the weights of the last layers, which are not subject to the regularization. When $W^{(L)}$ and $W^{(L-1)}$ are fixed, the regularization is less able to force the network's weights towards obtaining robustness, and at the same time the network is less able to perform its classification task with a good test accuracy. In addition, as typically the size of the last layers of the network becomes smaller and smaller towards the last layer (consisting of less neurons), applying Jacobian regularization on earlier layers is significantly more computationally demanding.

\section{Experimental results for the CIFAR-100 dataset}
\label{Experimental results for the CIFAR-100 dataset}
\begin{figure*}
  \centering
  \begin{subfigure}[b]{0.5\linewidth}%
  \centering
  \includegraphics[scale=0.17]{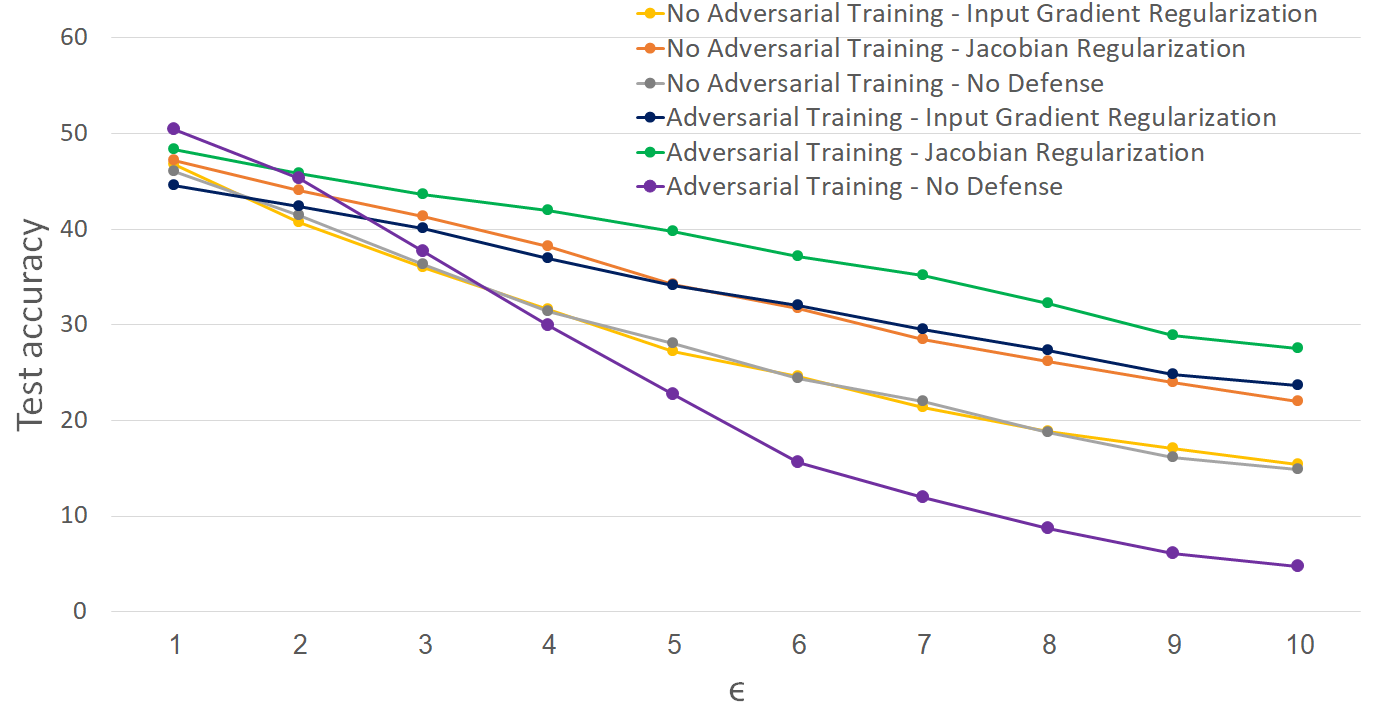}
  \caption{FGSM attack}
  \label{fig:CIFAR_100_FGSM}
  \end{subfigure}%
  \hfill
  \begin{subfigure}[b]{0.5\linewidth}%
  \centering
  \includegraphics[scale=0.17]{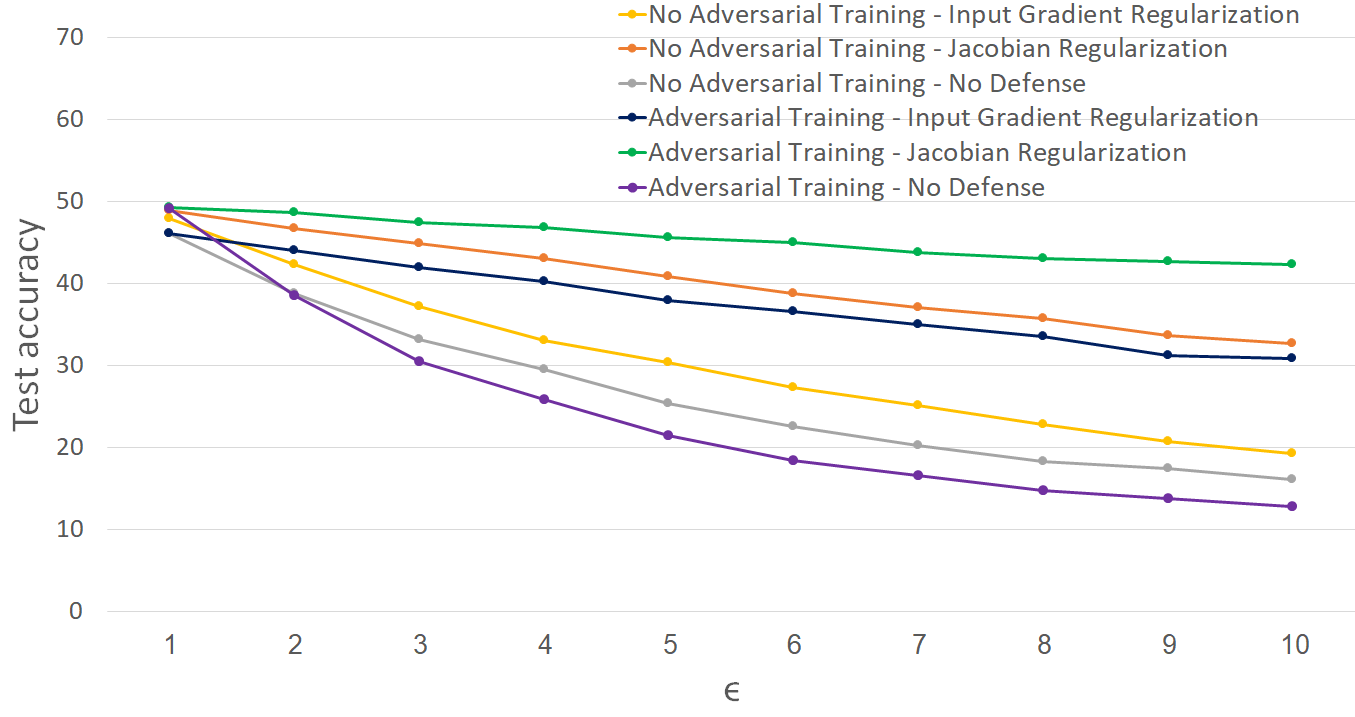}
  \caption{JSMA attack}
  \label{fig:CIFAR_100_JSMA}
  \end{subfigure}%
  \caption{CIFAR-100 test accuracy under FGSM (left) and JSMA (right) attacks for different values of $\epsilon$}
\label{fig:FGSM_JSMA_CIFAR_100}
\end{figure*}

For CIFAR-100 we used the exact same network as for CIFAR-10. As CIFAR-100 is a more computationally demanding dataset ($K=100$), training was done with a mini-batch size of 96 inputs. With this network we obtained a test accuracy of $59.64\%$. Training with Jacobian regularization was done with a weight of $\lambda = 0.02$, which we found to provide a good balance between the cross-entropy loss and the Jacobian regularization.
We found that the Cross-Lipschitz regularization is not scalable to large values of $K$ (unreasonable runtime), and it is therefore not included in the results.
Table~\ref{table:CIFAR-100_DeepFool_Results} presents the results for the DeepFool attack. Fig.~\ref{fig:FGSM_JSMA_CIFAR_100} shows the results for the FGSM and JSMA attacks.

\begin{table}
\centering
\scriptsize
\caption{Robustness to DeepFool attack for CIFAR-100}
\begin{tabular}{||c c c||} 
\hline
Defense method & Test accuracy & $\hat{\rho}_{adv}$ \\ [0.5ex] 
\hline\hline
No defense & $59.64\%$ & $0.61$ x ${10^{-2}}$ \\
Adversarial Training & $60.67\%$ & $1.23$ x ${10^{-2}}$ \\
Input Gradient regularization & $62.55\%$ & $0.76$ x ${10^{-2}}$ \\
Input Gradient regularization $\&$ Adversarial Training & $60.66\%$ & $1.45$ x ${10^{-2}}$ \\  
Jacobian regularization & $59.20\%$ & $1.55$ x ${10^{-2}}$ \\
 Jacobian regularization $\&$ Adversarial Training & $60.38\%$ & $3.34$ x ${10^{-2}}$ \\
 \hline
\end{tabular}
\label{table:CIFAR-100_DeepFool_Results}
\end{table}

Interestingly, adversarial training yields good robustness results under the FGSM and JSMA attacks for small magnitude attacks (small values of $\epsilon$), yet for stronger attacks adversarial training actually harms the network's robustness.

\section{Influence of the amount of adversarial examples in the training mini-batch}
\label{Influence of the amount of adversarial examples in the training mini-batch}

In this section, we analyze the effect of the amount of adversarial examples used in training on the robustness of the network. 
We found that incorporating different amounts of adversarial examples in the training mini-batch can lead to very different final network robustness results. Adding more perturbed inputs to the training set does not necessarily increase the network's robustness. The experiments depicted hereafter are done on the MNIST dataset, with the same network as described in the main paper and with the same mini-batch size of 500 inputs. We evaluate various  proportions between the number of original training examples and the number of perturbed adversarial examples. All the perturbed inputs were generated using the DeepFool attack method.

\begin{table}
\centering
\scriptsize
\caption{Robustness to DeepFool attack for MNIST with adversarial training, with different percentages of perturbed examples in a training mini-batch of 500 inputs}
\begin{tabular}{||c c c||} 
 \hline
 Defense method & Test accuracy & $\hat{\rho}_{adv}$ \\ [0.5ex] 
 \hline\hline
 No defense & $99.08\%$ & $20.67$ x ${10^{-2}}$ \\
 Adversarial training, $10\%$ perturbed & $98.99\%$ & $23.73$ x ${10^{-2}}$ \\ 
 Adversarial training, $20\%$ perturbed & $98.8\%$ & $22.29$ x ${10^{-2}}$ \\
 Adversarial training, $30\%$ perturbed & $98.91\%$ & $20.71$ x ${10^{-2}}$ \\
 Adversarial training, $40\%$ perturbed & $98.96\%$ & $19.53$ x ${10^{-2}}$ \\
 Adversarial training, $50\%$ perturbed & $98.72\%$ & $18.74$ x ${10^{-2}}$ \\
 \hline
\end{tabular}
\label{table:7}
\end{table}

\begin{table}
\centering
\scriptsize
\caption{Robustness to DeepFool attack for MNIST with Jacobian regularization ($\lambda = 0.1$) and adversarial training, with different percentages of perturbed examples in a training mini-batch of 500 inputs}
\begin{tabular}{||c c c||} 
 \hline
 Defense method & Test accuracy & $\hat{\rho}_{adv}$ \\ [0.5ex] 
 \hline\hline
  Jacobian regularization \& no adversarial training & $98.44\%$ & $34.24$ x ${10^{-2}}$ \\
 Jacobian regularization \& adversarial training $10\%$ perturbed & $98.06\%$ & $35.57$ x ${10^{-2}}$ \\ 
 Jacobian regularization \& adversarial training $20\%$ perturbed & $97.92\%$ & $37.14$ x ${10^{-2}}$ \\
  Jacobian regularization \& adversarial training $30\%$ perturbed & $98.07\%$ & $34.32$ x ${10^{-2}}$ \\
 Jacobian regularization \& adversarial training $40\%$ perturbed & $97.87\%$ & $34.43$ x ${10^{-2}}$ \\
 Jacobian regularization \& adversarial training $50\%$ perturbed & $98.03\%$ & $36.89$ x ${10^{-2}}$ \\
 \hline
\end{tabular}
\label{table:8}
\end{table}

The results are examined under the DeepFool attack and evaluated using the $\hat{\rho}_{adv}$ metric. Table~\ref{table:7} presents the results without any defense except adversarial training. Table~\ref{table:8} shows the results with both adversarial training and Jacobian regularization.
The results suggest that using more perturbed inputs in the training mini-batch does not necessarily improve the final robustness to adversarial attacks, but rather there is specific balance for which optimal robustness is achieved. It can also be observed that this balance changes with the addition of the Jacobian regularization to the adversarial training.

\section{Jacobian regularization - influence of the hyper-parameter $\lambda$}
\label{Jacobian regularization - influence of the hyper-parameter lambda}

The influence of variation in the value of the hyper-parameter $\lambda$, which balances the cross-entropy loss and the Jacobian regularization, is given in Table~\ref{table:10} for MNIST and Table~\ref{table:11} for CIFAR-10.

For MNIST, the obtained robustness is higher as the value of $\lambda$ increases, with a slight degradation in the test accuracy, up to the value of $\lambda=0.1$. For $\lambda > 0.1$, the obtained robustness declines along with a substantial degradation in the test accuracy.

\begin{table}
\centering
\scriptsize
\caption{Influence of $\lambda$ on Jacobian regularization for MNIST. The chosen value is in bold.}
\begin{tabular}{||c c c||} 
 \hline
 $\lambda$ & Test accuracy & $\hat{\rho}_{adv}$ \\ [0.5ex] 
 \hline\hline
0 & $99.08\%$ & $20.67$ x ${10^{-2}}$ \\  \hline
0.001 & $99.00\%$ & $22.21$ x ${10^{-2}}$ \\  \hline
0.005 & $98.96\%$ & $23.99$ x ${10^{-2}}$ \\  \hline
0.01 & $98.97\%$ & $26.34$ x ${10^{-2}}$ \\  \hline
0.02 & $98.76\%$ & $28.93$ x ${10^{-2}}$ \\  \hline
0.05 & $98.62\%$ & $32.39$ x ${10^{-2}}$ \\  \hline
\textbf{0.1} & $\textbf{98.44\%}$ & $\textbf{34.24}$ x $\mathbf{10^{-2}}$ \\  \hline
0.2 & $97.88\%$ & $34.18$ x ${10^{-2}}$ \\  \hline
0.5 & $96.92\%$ & $33.51$ x ${10^{-2}}$ \\  \hline
1 & $95.24\%$ & $33.48$ x ${10^{-2}}$ \\  \hline
\end{tabular}
\label{table:10}
\end{table}

\begin{table}
\centering
\scriptsize
\caption{Influence of $\lambda$ on Jacobian regularization for CIFAR-10.  The chosen value is in bold.}
\begin{tabular}{||c c c||} 
 \hline
 $\lambda$ & Test accuracy & $\hat{\rho}_{adv}$ \\ [0.5ex] 
 \hline\hline
0 & $88.79\%$ & $1.21$ x ${10^{-2}}$ \\  \hline
0.001 & $88.69\%$ & $1.45$ x ${10^{-2}}$ \\  \hline
0.005 & $88.65\%$ & $1.74$ x ${10^{-2}}$ \\  \hline
0.01 & $88.76\%$ & $1.80$ x ${10^{-2}}$ \\  \hline
0.05 & $88.73\%$ & $2.44$ x ${10^{-2}}$ \\  \hline
0.1 & $89.16\%$ & $2.69$ x ${10^{-2}}$ \\  \hline
\textbf{0.5} & $\textbf{89.16\%}$ & $\textbf{3.42}$ x $\mathbf{10^{-2}}$ \\ \hline
1 & $87.82\%$ & $4.08$ x ${10^{-2}}$ \\  \hline
2 & $86.30\%$ & $4.83$ x ${10^{-2}}$ \\  \hline
5 & $81.46\%$ & $4.60$ x ${10^{-2}}$ \\  \hline
\end{tabular}
\label{table:11}
\end{table}

Unlike MNIST, for CIFAR-10 there is a slight improvement in the test accuracy when Jacobian regularization is applied, even though for too large values of $\lambda$ the test accuracy deteriorates as well.

\section{Jacobian regularization and the JSMA attack}
\label{Jacobian regularization and the JSMA attack}

While our proposed Jacobian regularization outperforms all the other defense methods considered in the paper (adversarial training, Input Gradient regularization and Cross-Lipschitz regularization) under the DeepFool and the FGSM attacks, it gets inferior performance compared to the Input Gradient regularization technique for the JSMA attack on the MNIST dataset (for CIFAR-10 it is better). Here we provide a possible explanation for the reason that regularizing the Frobenius norm of the Jacobian does not perform as well for this type of attack. 

We start by a short description of the JSMA attack algorithm. This attack targets the $\ell_0$ pseudo-norm of the input image. According to a \emph{saliency map}, one pixel in the input image is changed in every epoch. This pixel is chosen such that the change in the classification output is maximal towards the choice of another target class. This attack is effective in minimizing visual detectability since a change in a small amount of pixels is less likely to be noticed by the human eye. 

To better understand the reason behind the failure of our method we rely on the mathematical analysis provided in the main paper. 
This analysis shows that the Jacobian regularization defense, based on the regularization of the Frobenius norm, aims at maximizing the minimal $\ell_2$ distance between the original input and a perturbed version of that input that would cause a misclassification. This provides an explanation to the behavior that we see under JSMA, as the JSMA attack changes a minimal amount of pixels and the Jacobian regularization aims at maximizing the $\ell_2$ distance.

Better robustness to the JSMA attack would be achieved by maximizing the $\ell_0$ distance to adversarial examples with the minimal $\ell_0$ distance from the original input, which is a different goal than maximizing the $\ell_2$ distance. As we shall see now, these two goals are not necessarily aligned.
Let us assume that $n$ pixels are changed in the input image, each changed by a value of $A$. The $\ell_0$ pseudo-norm distance between the original image and the perturbed version is $n$, as only $n$ pixels were changed. However, the $\ell_2$ distance between the two is $\sqrt{nA^2} = \sqrt{n}|A|$.
For this reason, an $\ell_2$-based defense method penalizes large values of $|A|$ much more than large values of $n$, whereas an $\ell_0$ attack disregards the value of $|A|$ and only aims at minimizing $n$.

From this we draw the conclusion that an $\ell_2$-based defense is more effective against attacks that make small changes to a large number of pixels such as DeepFool or FGSM, than against attacks that make large changes to a small amount of pixels, which is the case of the JSMA attack.

\bibliographystyle{splncs}
\bibliography{egbib}

\end{document}